\documentclass[runningheads]{llncs}
\usepackage{graphicx}
\usepackage[T1]{fontenc}
\usepackage{bbm}
\usepackage{microtype}
\usepackage[breaklinks]{hyperref}

\begin{document}
\title{Scalable handwritten text recognition system for lexicographic sources of under-resourced languages and alphabets}
\titlerunning{Scalable HTR system for lexicographic sources}
% If the paper title is too long for the running head, you can set an abbreviated paper title here
%
\author{Jan Idziak\inst{1} \and
Artjoms Šeļa\inst{2}%\orcidID{1111-2222-3333-4444}
\and
Michał Woźniak\inst{2}%\orcidID{2222--3333-4444-5555}
\and
Albert Leśniak\inst{2}
\and
Joanna Byszuk\inst{2}
\and
Maciej Eder\inst{2}%\orcidID{0000-0002-1429-5036}
}
\authorrunning{J. Idziak et al.}
% First names are abbreviated in the running head.
% If there are more than two authors, 'et al.' is used.
%
\institute{independent scholar, Singapore\\
\email{jan.idziak@gmail.com}
\and
Institute of Polish Language, Polish Academy of Sciences,\\ al. Mickiewicza 31, 31-120 Krakow, Poland\\
\email{\{artjoms.sela,michal.wozniak,albert.lesniak,\\ joanna.byszuk,maciej.eder\}@ijp.pan.pl}}
\maketitle   % typeset the header of the contribution

\begin{abstract}

The paper discusses an approach to decipher large collections of handwritten index cards of historical dictionaries. Our study provides a working solution that reads the cards, and links their lemmas to a searchable list of dictionary entries, for a large historical dictionary entitled the \textit{Dictionary of the 17\textsuperscript{th}- and 18\textsuperscript{th}-century Polish}, which comprizes 2.8 million index cards. We apply a tailored handwritten text recognition (HTR) solution that involves (1) an optimized detection model; (2) a recognition model to decipher the handwritten content, designed as a spatial transformer network (STN) followed by convolutional neural network (RCNN) with a connectionist temporal classification layer (CTC), trained using a synthetic set of 500,000 generated Polish words of different length; (3) a post-processing step using constrained Word Beam Search (WBC): the predictions were matched against a list of dictionary entries known in advance. Our model achieved the accuracy of 0.881 on the word level, which outperforms the base RCNN model. Within this study we produced a set of 20,000 manually annotated index cards that can be used for future benchmarks and transfer learning HTR applications.

\end{abstract}
\section{Introduction}

Decades of lexicographic work that was done before the popularization of machine-readable texts provide rich lexicographic and/or linguistic data that is extremely hard to reuse today or to be integrated into modern databases, corpora and collections. Not only are these original resources handwritten, but they are also unstructured, or at best their structure is limited to an alphabetical order of the respective items. Card files served as tools of lexicographic description and, when collected into catalogues, allowed random access to vast bodies of lexical information. These cards were building blocks of lexicons and dictionaries, long before corpus linguistics that relied on digitized texts appeared \cite{landau_dictionaries_2001}. The lexicographic resources in question involve millions of handwritten cards for various historical dictionaries, ranging from Latin (with the archetypical \textit{Thesaurus Linguae Latinae}, one of the first initiatives of this kind), to medieval and modern language varieties (Middle Dutch, Old Czech, Old Norse Prose, or Middle High German to name but a few). In most cases, the index cards are acquired throughout several decades – sometimes dating back to the 19\textsuperscript{th} century – and they contain comprehensive documentation for all known words of respective language varieties. 

The lexicographic collections held at the Institute of Polish Language of the Polish Academy of Sciences are no exception, with its extensive card catalogues of \textit{The Old Polish Dictionary}, \textit{The Dictionary of the 17\textsuperscript{th}- and 18\textsuperscript{th}-century Polish}, \textit{The Dictionary of Polish Dialects}, \textit{The Great Dictionary of Polish}, as well as a few onomastic dictionaries of proper nouns. A single index card contains a lemma (a base word in a header) followed by its context excerpted from actual historical documents. Stored in dedicated boxes and alphabetized, the index cards are used by lexicographers to compose subsequent dictionary entries. Since most of the aforementioned dictionaries are not completed yet, the handwritten index cards serve as a work-in-progress source of information, and, even in the case of the dictionaries that are already published, the index cards are still valid as their supplementary materials. The biggest challenge, however, is that they are not machine-readable, and not linked to the searchable lexicographic databases. 

While recent years saw a development of numerous approaches to optical and handwritten text recognition (HTR) also in the relation to humanistic data, e.g. Transkribus \cite{kahle_transkribus_2007,muehlberger_transforming_2019} which provides excellent performance, such solutions are better suited to longer texts, fewer scribal hands and require significant amounts of training data \cite{franzini_attributing_2018}, also posing limitations as to the number of pages that can be annotated. Meanwhile, the index cards contain short excerpts, typically no more than a sentence of context next to the lemma, followed by source description, and are produced by numerous lexicographers, often showing inconsistent handwriting style that can be understood only by themselves or other team members. In fact, while the second poses a significant challenge, also in the case of the need to prepare manually annotated training set, computer vision methods which rely on less easy to observe patterns hold great promise of perhaps outperforming human reading of more illegible scribblings. 

This project sets up an operational workflow for retrieving lexical data from handwritten card catalogues, followed by matching their lemmas to the list of dictionary entries. To build and test a prototype, we have chosen \textit{The Dictionary of the 17\textsuperscript{th}- and 18\textsuperscript{th}-century Polish}, which is a good example of a lexicographic source that combines traditional materials of 2.8 million handwritten index cards with modern technologies – the dictionary itself is a fully digital database, and it is linked to an annotated corpus of the 17\textsuperscript{th}-century Polish \cite{ogrodniczuk_connecting_2019,bronikowska_electronic_2020}. Moreover, in a pilot study a small selection of the index cards was manually mapped onto a list of dictionary entries \cite{bilinska-brynk_paper_2020}. This preliminary work, however, clearly shows that manual mapping is hardly feasible in real-scale setups and would involve an immense effort expressed in thousands of working hours. Our project aims at overcoming this limitation by using an automated approach that would simplify the work of lexicographers in preparing the digital entries. The main goal of the project, however, goes beyond the prototype applied to the 17\textsuperscript{th}-century lexicographic sources. The diverse range of the obtained outcomes makes this study potentially interesting both from a computer sciences point of view, as well as from a digital humanities perspective.

The research presented in this paper provides the following contributions:

\begin{enumerate}

\item
We propose a unified modular workflow that is adjustable to any language, since the model relies solely on a synthetic dataset; our workflow can be easily extended to other under-resourced languages, including languages with extended Latin alphabets or non-Latin scripts.

\item
We provide a working HTR detection and recognition prototype (also as a deployed demo web application) that outperforms baseline models significantly.

\item
We provide a synthetic dataset of artificially generated 500,000 words in Polish, supplemented by another set of 30,000 random strings with uniform distribution of Polish diacritics.

\item
We offer a manually labelled set of 20,000 words in Polish to be used as ground truth in future applications and model evaluation settings. 

\end{enumerate}

\section{Related work}

Modern HTR heavily leans towards solutions based on Artificial Neural Networks and it was shown that various architectures of deep learning improve performance in the handwriting recognition task \cite{graves_offline_2008,pal_handwritten_2010,voigtlaender_handwriting_2016,gupta_synthetic_2016-1,jaderberg_spatial_2016,graves_multi-dimensional_2007,doetsch_fast_2014,yin_robust_2014,xiao_deep_2019} compared to Hidden Markov models \cite{sanchez_icfhr2014_2014,sanchez_icfhr2016_2016}. Handwritten text imposes severe challenges for a machine vision technologies that we counter by combining several known methods: (1) the problem of significant variation in writing style, shapes, sizes and possible deformations of characters is solved through spatial deformation rectification achieved with the Spatial Transformation Network that learns translation-invariant features \cite{jaderberg_spatial_2016}. (2) Since source material in HTR tasks is often based on a large and diverse lexicon which adds to the difficulty of decision making of the model, Word Beam Search is often used to decode CTC layer and constrain prediction errors \cite{scheidl_word_2018}. Many best-performing post-OCR error correction systems depend on contextual awareness provided by a language model \cite{rigaud_icdar_2019}, which cannot be adapted to our case of isolated index words. Instead we rely on lexicon-aided decoding of the model’s predictions that broadly follows the character-level error correction framework \cite{farra_generalized_2014}. (3) It is often very difficult to achieve a system generalizability in HTR in a specific domain because of the lack of a large amount of ground truth handwriting samples. Recent studies propose to compensate this by generating vast amounts of synthetic images from character strings \cite{jaderberg_synthetic_2014}. It was even shown that CNN algorithm trained only on the synthetic data outperformed other methods on the text detection task \cite{gupta_synthetic_2016-1}. (4) Finally, the level at which text segmentation for HTR is deployed was subjected to discussion. Early approaches \cite{marti_iam-database_2002-1} employed line segmentation in the HTR context, while word segmentation or character region awareness methods \cite{gupta_synthetic_2016-1,he_deep_2015} are gaining more popularity recently.

\section{Data}

\subsection{Original data}

Our main focus was the \textit{The Dictionary of the 17\textsuperscript{th}- and 18\textsuperscript{th}-century Polish} (\url{https://sxvii.pl/}), a partially-completed lexicographic database that is based on roughly 2.8 million index cards manually filled by different hands (in rare cases typewritten) in the years 1954–1995 \cite{bronikowska_electronic_2020,bilinska-brynk_paper_2020}. The cards are stored in 836 alphabetized boxes, and, after a digitization project conducted in the years 2010–2015, they are now freely accessible in bitmap image format, saved under names linking to their respective boxes (e.g. \textit{Egzekucja}–\textit{Ekspediowanie}). For the sake of this study, we drew a sample of 100,000 cards taken uniformly across all boxes, which made up our primary dataset (denoted as \texttt{PL-100k-main} hereafter).

\begin{figure}
  \begin{center}
  \includegraphics[width=0.35\columnwidth]{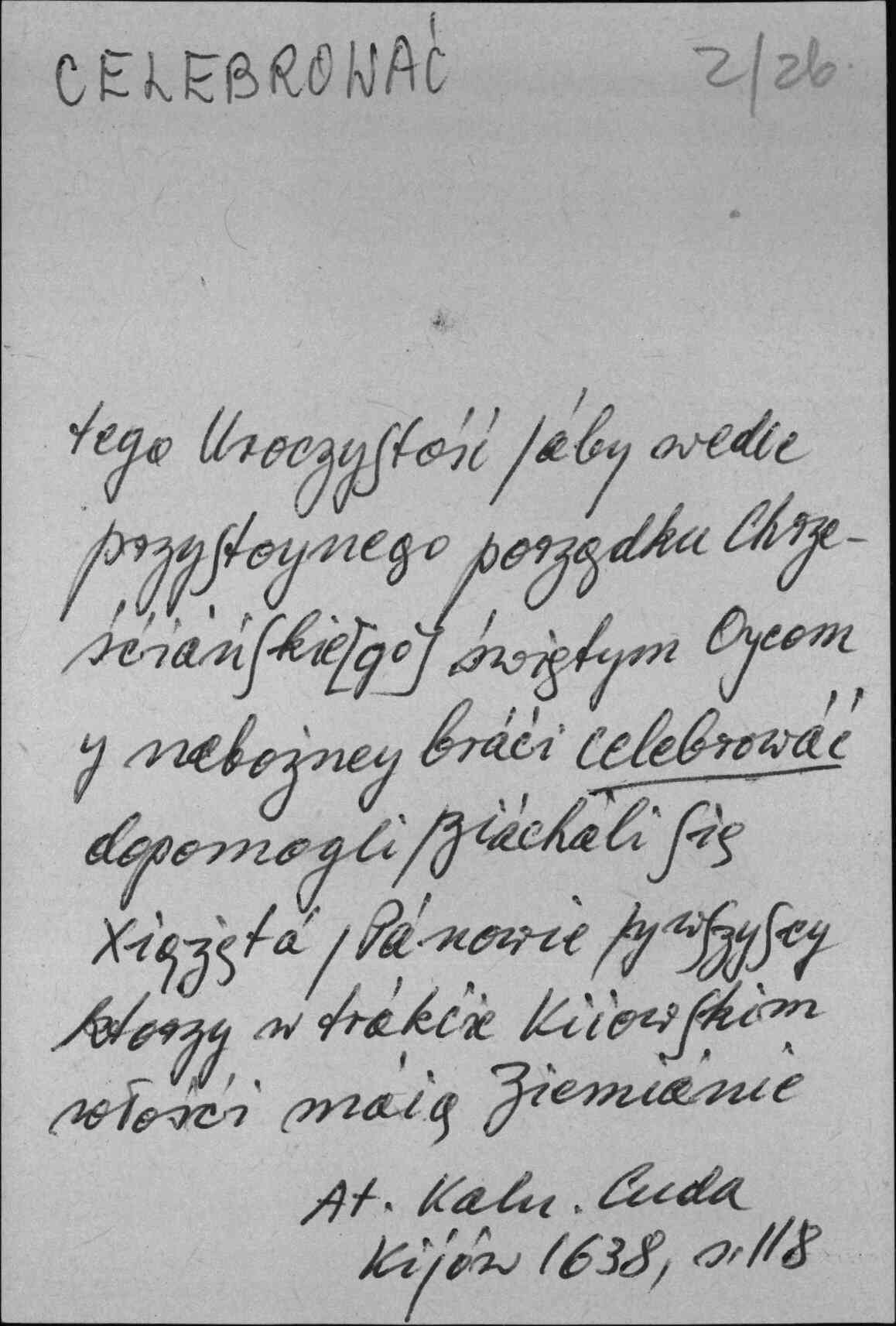}
  \caption{A handwritten index card from \textit{The Dictionary of the 17\textsuperscript{th}- and 18\textsuperscript{th}-century Polish}.}
  \end{center}
\end{figure}

Almost all of the cards – rare exceptions being phonetic variant pointers – followed a conventional scheme for encoding a given word’s attestation (see Fig. 1):  (1) a lemma served as an index which ensured navigation and accessibility, while (2) the body of the card registered immediate context of the word occurrence, followed by (3) a bibliographical reference to sources and sometimes including other information (grammatical form, alternate spelling, etc.). Our lemma detection procedure at times encountered difficulties when the additional information about grammatical form was written next to the header lemma, rather than within the card’s body. A significant part of the cards from the \textit{The Dictionary of the 17\textsuperscript{th}- and 18\textsuperscript{th}-century Polish} had their index word written in “handwritten capitals” (majuscule alphabet) to keep the words recognizable, which limits the variability of shapes and maintains somewhat clear boundaries between letters. A mix of minuscule and all-majuscule handwriting meant that the HTR model should be able to perform simultaneously on both of these modes of writing, which added an additional step to the transfer learning workflow (as discussed below).  

The list of all the 86,000 dictionary entries of the \textit{The Dictionary of the 17\textsuperscript{th}- and 18\textsuperscript{th}-century Polish} is publicly accessible (\url{https://sxvii.pl/}). We used the list to serve as a constrained set of possible words to match them against the resulting predictions from our HTR system.

\subsection{Synthetic data for training}

The majority of already existing training datasets for HTR are limited to English. Since Polish language uses an extended Latin script with additional 16 diacritics (8 lowercase and 8 capitalized), we had to make sure that Polish is at least partially represented \cite{grzelak_analyze_2019}. Our transfer-learning approach involved the already existing large dataset of English words CVIT \cite{krishnan_matching_2016} that we further enhanced with an artificially generated set of Polish words. To this end, we randomly excerpted 500,000 actual Polish words from the corpus of 17\textsuperscript{th}-century Polish texts \textit{Korba} \cite{ogrodniczuk_connecting_2019}, and generated their bitmap representations using a variety of fonts that mimicked the handwriting both in lowercase and uppercase (this is our \texttt{PL-500k-synthetic} dataset). Not only were the words and the font shapes picked at random, but also the final bitmap images were slightly distorted using different augmentation steps. The augmentation distortion included (1) posterization – which maximizes the image contrast, (2) equalization of the image histogram, (3) solarization – which inverts all pixel values above a threshold, (4) affine geometric transformation. Consequently, the set we obtained consisted of artificial bitmap representations of actual Polish words. This allowed for training the representation of the Polish diacritics, while the proportion of diacritics to standard characters followed their natural distribution.

Apart from the above \texttt{PL-500k-synthetic} dataset, we also prepared an additional set of 30,000 randomly generated strings containing solely Polish diacritical marks – with uniform distribution – in order to improve performance for rare polish diacritics. We denote this set as \texttt{PL-30k-diacritics}.

\subsection{Manually labelled subsets}

To obtain a carefully curated set for evaluation purposes, we drew a small sample from the original dataset, namely 20,000 bitmap images, that were then manually corrected for their bounding boxes, and manually labeled by the project members. This set (\texttt{PL-20k-hand-labelled}) was our primary evaluation set, since it provided ground truth for both neatly cropped images and corrected labels. The set has been made publicly available for future benchmarking and applications. In order to facilitate the tedious annotation work, we applied the annotation tool Prodigy (\url{https://prodi.gy/}). 

Yet another manually annotated subset was prepared (\texttt{PL-3k-boundaries}) to optimize the performance of word detection. We drew at random 3,000 images from the original dataset, applied the word detection module as discussed above, and manually-corrected the resulting bounding boxes around the detected lemmas.

\begin{figure*}
  \includegraphics[width=\textwidth]{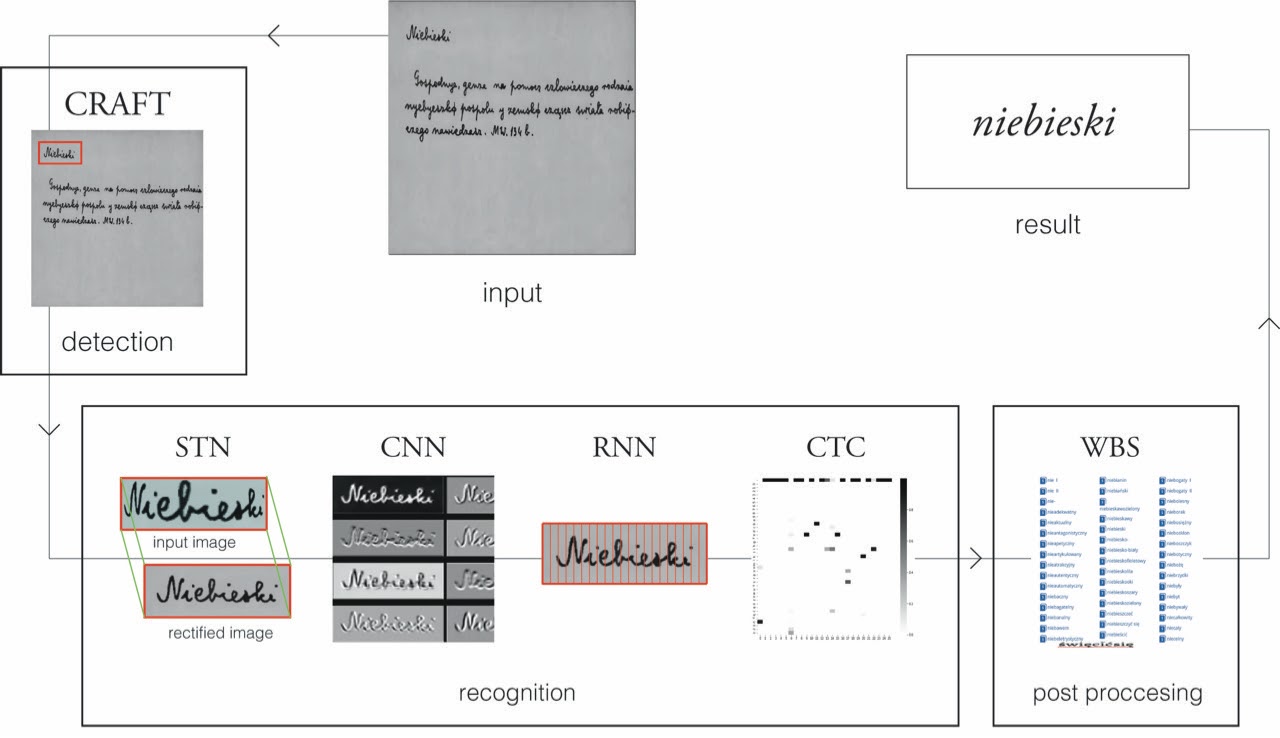}
  \caption{The HTR workflow, including the detection, recognition, and post-processing steps.}
\end{figure*}

\section{Methodological workflow}

Our general workflow involved three main parts: (1) detection, (2) recognition, and (3) postprocessing. The first two steps are based on deep learning and convolutions, while the third step is an optimization technique aimed at improving the final predictions of the neural network.

\subsection{Detection of the index word}

Since all cards had a header lemma, usually placed distinctly from the body text, it is relatively easy to access it: we cut a card to top 300 pixels in order to optimize for computation time, and used Keras OCR Craft model for text detection and bounding box assignment \cite{baek_character_2019}. Then for each card, we identified the first bounding box located in the top-left area, which in the vast majority of cases contains the word in question. We selected 3,000 hand-labeled images to optimize the position of the bounding boxes (the \texttt{PL-3k-boundaries} set).

\subsection{Recognition}

The recognition stage can be broken down to four consecutive components based on TPS-ResNet-BiLSTM-CTC architecture proposed by Baek at al. \cite{baek_what_2019}. We used already existing pre-trained model that we further improved:
 
\begin{enumerate}
\item
Input text image was first rectified with the help of the Spatial Transformer Network, or STN  \cite{jaderberg_spatial_2016} with Thin Plate Spline (TPS) transformation \cite{yin_robust_2014}.  The aim of this step is to ensure that the images are consistent in terms of contrast, saturation and so forth, and thus easier to process at the feature extraction step.

\item
Feature extraction using a Convolutional Neural Network (CNN) setup. It extracted relevant features from the image and focused on attributes that are characteristic to particular characters. After a few preliminary tests, we chose a ResNet backbone, because it provided a clear improvement in accuracy.

\item
The features extracted in step 2 are fed sequentially into the Bidirectional LSTM layer (BiLSTM).

\item
Finally, the Connectionist Temporal Classification (CTC) layer was involved. The benefit of using it is at least two-fold. Firstly, the predictions show the ability to overcome a variable size of the input sequence, even if the number of features is fixed. Secondly, the CTC layer accounts for words with repeated letters, thus helping to differentiate between, say, the words “to” and “too”. Also, because most of the publicly available datasets used English, a standard Attention layer would not guarantee sufficiently good performance for other languages. The CTC layer, on the contrary, provided a matrix of all possible predictions which could be further generalized beyond English. 
\end{enumerate}

\begin{figure}
  \begin{center}
  \includegraphics[width=0.35\columnwidth]{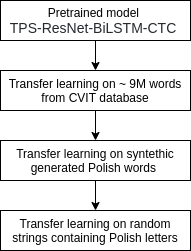}
  \caption{Transfer learning workflow.}
  \end{center}
\end{figure}

\subsection{Transfer learning}

In our approach we first took an existing model pre-trained on two datasets: MJSynth (MJ) \cite{jaderberg_synthetic_2014} and SynthText (ST) \cite{jaderberg_spatial_2016}. These datasets are not designed for HTR problems and do not provide Polish characters, therefore we involved a few additional datasets to enhance the model. Firstly, we added the CVIT database with 9 million images of handwritten words based on English corpus. Subsequently, further domain improvement was done using Polish words from a synthetic dataset \texttt{PL-500k-synthetic} as discussed above. The last step involved enhancing the performance specifically for the recognition of Polish diacritics. To achieve this, we used the randomly generated strings containing all characters with uniform distribution (\texttt{PL-30k-diacritics}).

Initially, we also used the IAM handwritten database containing carefully-labelled handwritten words \cite{marti_iam-database_2002-1}. After several rounds of transfer learning tests, however, it became clear that the dataset representation and the sample images significantly differ from texts produced in natural conditions. This is caused by the fact that the words in the IAM handwritten database are cut closely around the word outlines, which confuses the recognition system. Our observations suggest that using this dataset leads to overfitting of neural network models with a large number of parameters. Consequently, any applications based on word representation of the IAM handwritten dataset do not seem suitable in real life situations. We would even argue that this dataset should not be used as the main dataset for performance evaluation as well.

\subsection{Postprocessing}

The predictions as produced by our workflow will always contain some wrongly reconstructed words. However, some of these mistakes are relatively easy to correct, due to the non-random nature of the language. For instance, a human would instinctively guess that the string “cvolution” resembles the word “evolution”. The string “ancl” would require an additional split second to decipher, because it could be reconstructed as “and”, as “ant” and perhaps even as “uncle” or “anele”. Since the list of possible words is limited, an optimization algorithm can be used that matches the input sequence with the nearest element from the closed set of words known in advance. Real-life situations might be more challenging, given the fact that new words emerge at times, e.g. the algorithm would disregard the string “covfefe” as a valid word, and would replace it either with “coffee” or with “coverage”. 

In the case of our project, however, a vast majority of words written down on index cards match the close set of words stored as a list of 86,000 dictionary entries. Additionally, we took advantage of the fact that the index cards are alphabetized (as is the list of dictionary entries), and stored in 836 boxes with known ranges of the alphabet ascribed to each box. Consequently, in our constrained Word Beam Search (WBC) approach not only did we take into consideration the CTC match of the predictions and the expected dictionary entries, but we also assumed that it is very unlikely to match an index card from a given box to a word belonging to a distant part of the alphabetized list.

In the first round of our procedure, we additionally involved a quality check step. To this end, we prepared the aforementioned \texttt{PL-20k-hand-labelled} dataset: we randomly selected 20,000 index cards, manually checked the predictions, and corrected all the misrecognized words. The aim of this step was two-fold: firstly, to allow for re-training the model with the clean ground-truth dataset (while keeping in mind that for the sake of this study we didn’t apply it, for the sake of a clear-cut separation of the training set and the evaluation set) and secondly, to prepare 20,000 manually corrected cards for public access for future benchmarking and improving HTR models.

\begin{figure*}
  \includegraphics[width=\textwidth]{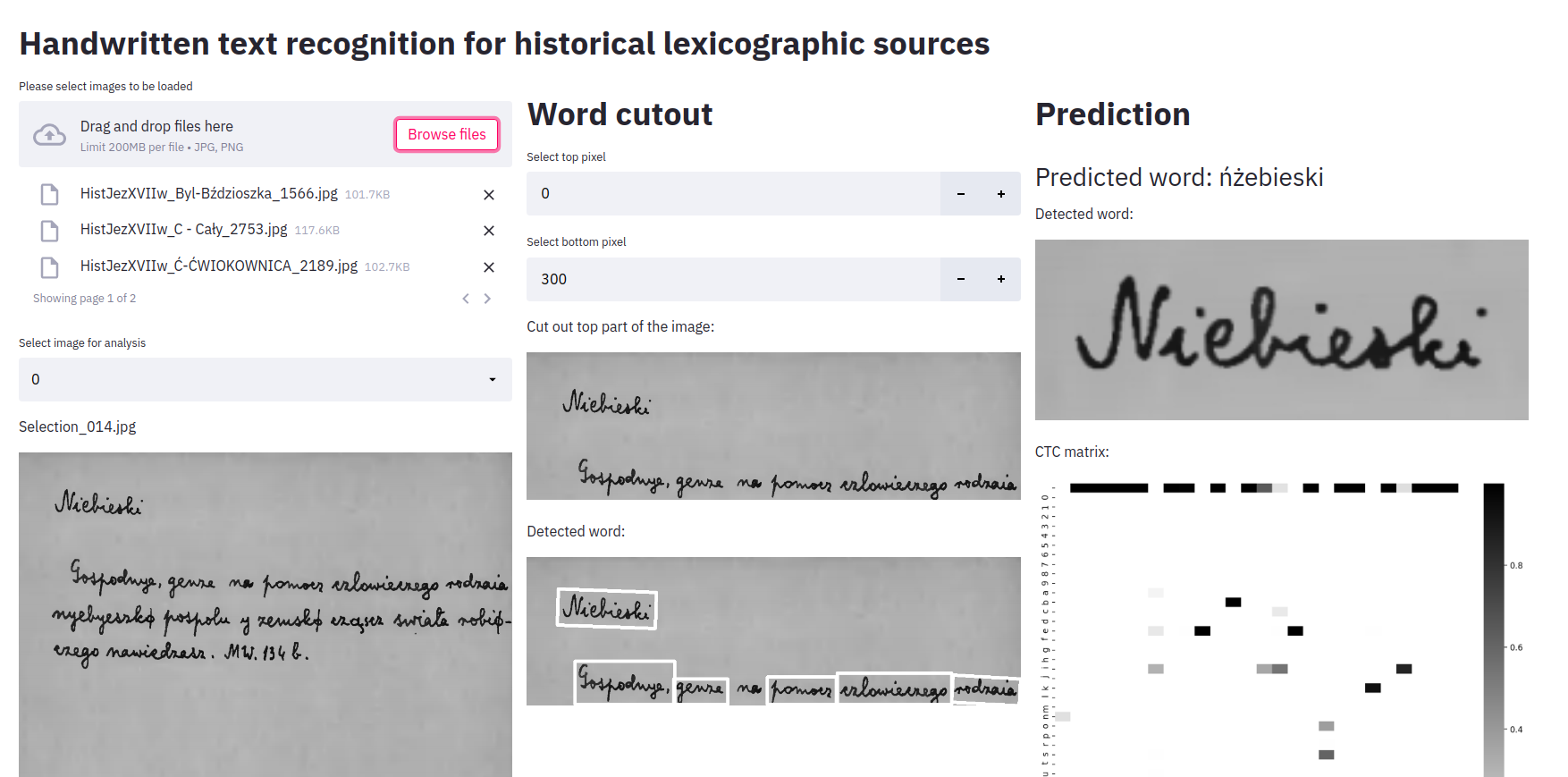}
  \caption{Online demo application of the HTR system.}
\end{figure*}

\section{Application}

Alongside our full-size system that is capable of consecutively processing millions of cards, we also designed an online application for testing our workflow and visually presenting how a given index card is being processed. After the file is selected, the app automatically performs cutting, detection and recognition of the word. In post-processing, the app optimizes for the best path CTC encoding. Both vanilla prediction and the CTC matrix are available for inspection.

\section{Results}

The results of the system are based on accuracy scores achieved on a subset of the original data (the \texttt{PL-100k-main} datset). The detection component of the workflow achieved 0.93 of intersection over union on 3,000 hand labeled images (the \texttt{PL-3k-boundries} dataset). The results for the recognition model are best presented in comparison to a baseline model. We have excluded 20,000 index cards (the \texttt{PL-20k-manual}) to serve as a test set, the average word length was 5.99 characters. Accuracy is calculated as the number of words that were classified correctly divided by the number of all words in the dataset:

$$ Acc = \frac{1}{n} \sum_{i}^{n}\mathbbm{1}_{y_{i} = \hat{y}_{i}}$$

We use Levenshtein distance as the edit distance. Average edit distance is calculated as the sum of the edit distance divided by the number of words. Average normalized edit distance is calculated as averaged edit distance per number of characters in the word:

$$ \Delta_{Lnorm}(i) = \frac{\Delta_{L}(y_{i}, \hat{y}_{i})}{\max(\mathrm{length}(y_{i}), \mathrm{length}(\hat{y}_{i}))} $$

We compared the performance of three decoding approaches of the CTC layer in the RCNN model: (1) best path decoding, (2) word beam search, and (3) constrained word beam search. The results are shown in Table 1. As can be observed, we achieved a significant improvement over the base model offered by an RCNN (TPS-ResNet-BiLSTM-CTC) model with no transfer learning and no CTC and Word Beam Search refinement. 

\begin{table*}
  \caption{Results achieved by the HTR models. All of them are based on TPS-ResNet-BiLSTM-CTC architecture. BP – best path, WBS – word beam search, WBS-C – constrained word beam search, POL – model trained on the Polish synthetic set.}
\begin{tabular}{llccc}
\hline
~ & Word & Normalised & Edit & Average edit \\
Model & accuracy~~~~~ & ~~~edit distance~~~ & ~~~~~distance~~~~~ & ~~on misclassified~~ \\
\hline
BP & 0.3755 & 0.1898 & 1.0871 & \textbf{1.7484} \\
WBS & 0.0995 & 0.6194 & 6.1763 & 6.8711  \\
BP-POL & 0.4332 & 0.2246 & 1.2120 & 2.0945 \\
WBS-POL & 0.6655 & 0.2033 & 1.0993 & 2.7063 \\
WBS-C-POL~~~~~~ & \textbf{0.8810} & \textbf{0.0479} & \textbf{0.3165} & 3.2125  \\
\hline
\end{tabular}
\end{table*}

Since our pipeline is, among other things, aimed to aid lexicographers in linking index cards to existing databases, the model that minimizes edit distance on \textit{wrong} predictions sometimes could be more useful than the WBS model that is designed for word-level accuracy and makes use of external information (alphabet range of a box). Average edit distance on wrongly recognized words (Table 1) shows this effect: both best path decoding models have less edits than WBS decoding (1.75 and 2.09), suggesting that in a realistic setting of manual work with the full-scale collection, less greedy models could provide more “useful” predictions despite the drop in word-level accuracy. In addition, the best-performing constrained WBS model has a higher rate of edits in misclassified words than unconstrained WBS which implies that sometimes words fall out of the box’s alphabetical range but then are still forcefully fitted to that range by WBS-C. Most probably, these  mistakes happen because index word position could be recognized incorrectly in a situation when a card has multiple words on the top.
\looseness-1

\section{Discussion}

Our results show a trade-off between a straightforward word-level accuracy and amount of noise captured by a model. High values of average edit distance on the misclassified words for the Word Beam Search models show that the algorithm is likely to pick up noise in the activation map and predict long words even though the best path encoding would ignore such activation. If the label is “a” and the predicted word is “aproksymacja”, the edit distance would be eleven. This is also visible for the vanilla base model with the Word Beam Search encoding: it achieves the accuracy of less than 0.1, while the base model is able to get to 0.37. Our best-performing model achieves the word-level accuracy of 0.88 due to highly constrained output, limited by the alphabetical range of a box for a source word, but at the same time its tendency to aggressively fit predictions to longer or unrelated words remains an issue.

The vanilla RCNN model had lower edit distances than the vanilla model with the knowledge of the full Polish alphabet. This happened because Polish diacritics extend possibilities of decoding and are simultaneously not frequently encountered. Thus, a model that is aware of Polish diacritics, could predict \textit{a} as \textit{ą} or \textit{e} as \textit{ę}, while in real life \textit{ę} and \textit{ą} are infrequent. At the same time, the vanilla base model would not make these errors as it does not have \textit{ą} or \textit{ę} available for a prediction at all.

Our work highlights the importance and possibilities of automated information extraction from a historical archive, on the example of index card catalogues of dictionaries of historical variants of Polish developed in the Institute of Polish Language of the Polish Academy of Sciences. The way in which index cards are organized, facilitates the recovery of some parts of the source structure (e.g. index lemma, body, references) and invites further processing, such as knowledge linking, that goes beyond the undiscriminating plain text recognition from a given image. That recognition of already structured information could be further extended with additional layers of layout analysis of a card image.

\section{Conclusion}

In this paper, we presented a HTR solution tailored for processing large collections of handwritten index card catalogues. Although primarily designed to deal with the Polish language and lexicographic sources, our solution also expands HTR applicability to under-resourced languages and alphabets, providing domain-specific dataset that could be further reused for a wide array of tasks. The linguistic archives around the world present a wide variety of historical, dialectal, onomastic and other lexicographic data. Such linguistic projects were often undertaken decades before the creation of corpus linguistic tools and digital databases. HTR pipelines could contribute greatly to quickening the pace of work on turning index cards into dictionaries. In the future, a similar approach could be used to globally solve the problem of linking data of the past to the existing resources and linguistic platforms which today massively inhabit the digital space and utilize its affordances. The recognition of different handwriting and definition building patterns can also serve as an invaluable resource for examining the conventions of work in dictionary project as well as the study of individual contributions.

\section*{Software and data}

Code, models and data: \url{https://github.com/perechen/htr_lexicography} Web demo application: \url{http://149.156.30.114:8503/}

The elements of the system implemented in (or inspired by) other solutions: synthetic data generation \cite{text_renderer}, detection model \cite{baek_character_2019}, recognition model \cite{baek_what_2019}, CTC and Word Beam Search \cite{scheidl_word_2018}.

The 17th-century index cards (the original dataset):
\url{https://rcin.org.pl/dlibra/publication/20029}.

\section*{Acknowledgements}

This research was partly conducted as a result of a project supported by Poland’s National Science Centre (project number UMO-2013/11/B/HS2/02795). The authors are grateful to Bartłomiej Borek for his IT support, which included setting up the server and the environment for conducting our experiments.

\bibliographystyle{splncs04}
\bibliography{biblio}

\end{document}